\documentclass[10pt,twocolumn,letterpaper]{article}

\usepackage{iccv}
\usepackage{times}
\usepackage{epsfig}
\usepackage{graphicx}
\usepackage{amsmath}
\usepackage{amssymb}
\usepackage{booktabs}

\usepackage{graphicx}
\usepackage{amsmath}
\usepackage{amssymb}
\usepackage{booktabs}

\usepackage{multirow}
\usepackage{makecell}
\usepackage{xspace}
\usepackage{pifont}
\usepackage[dvipsnames]{xcolor}
\usepackage{color, colortbl}

\usepackage{threeparttable}
\usepackage{enumitem}
\usepackage{arydshln}
\usepackage{booktabs}
\usepackage{multirow}
\usepackage{makecell}
\usepackage[normalem]{ulem}
\usepackage{colortbl}
\usepackage{subfloat}
\usepackage{subfigure}
\usepackage[ruled,vlined]{algorithm2e}
\usepackage{footnote}
\usepackage[accsupp]{axessibility} 

\SetKwComment{Comment}{$\triangleright$}{}

\definecolor{Gray}{gray}{0.8}
\definecolor{shallowGray}{gray}{0.9}
\definecolor{LightCyan}{rgb}{0.88,1,1}
\definecolor{Olive_Green}{rgb}{0.0, 0.55, 0.0}
\definecolor{a_color}{rgb}{0.0, 0.55, 0.0}
\definecolor{YellowGreen}{rgb}{0.60, 0.80, 0.20}
\definecolor{a_color}{RGB}{62,129,183}
\definecolor{ref_color}{RGB}{0, 255, 0}

\newcommand{\kanfix}[1]{{#1}}

\newcommand\blfootnote[1]{%
\begingroup 
\renewcommand\thefootnote{}\footnote{#1}%
\addtocounter{footnote}{-1}%
\endgroup 
}

\definecolor{Klein_Blue}{rgb}{0.0, 0.129, 0.6}
\usepackage[pagebackref,breaklinks,colorlinks]{hyperref}
\hypersetup{
    colorlinks=true,
    linkcolor=magenta,
    urlcolor=magenta,
    citecolor=Klein_Blue
    }

\iccvfinalcopy 

\def\iccvPaperID{6244}

\ificcvfinal\fi

\newlength\savewidth

\begin{document}

\title{TinyCLIP: CLIP Distillation via Affinity Mimicking and Weight Inheritance}

\author{Kan Wu$^{1,*}$, Houwen Peng$^{3,*,^\dagger}$, Zhenghong Zhou$^{2,*}$,  
Bin Xiao$^{3}$, Mengchen Liu$^{3}$, 
Lu Yuan$^{3}$, \\
Hong Xuan$^{3}$, Michael Valenzuela$^{3}$, Xi (Stephen) Chen$^{3}$, Xinggang Wang$^{2}$, Hongyang Chao$^{1}$, Han Hu$^{3}$ \\
$^1$ Sun Yat-sen University, $^2$ Huazhong University of Science \& Technology, $^3$ Microsoft
}

\maketitle
\ificcvfinal\thispagestyle{empty}\fi

\blfootnote{$^*$ Equal contribution. Kan and Zhenghong were interns of Microsoft.}

\blfootnote{$^\dagger$ Corresponding: houwen.peng@microsoft.com
}

\vspace{-6mm}
\begin{abstract}

In this paper, we propose a novel cross-modal distillation method, called TinyCLIP, for large-scale language-image pre-trained models. The method introduces two core techniques: affinity mimicking and weight inheritance. Affinity mimicking explores the interaction between modalities during distillation, enabling student models to mimic teachers' behavior of learning cross-modal feature alignment in a visual-linguistic affinity space. Weight inheritance transmits the pre-trained weights from the teacher models to their student counterparts to improve distillation efficiency. Moreover, we extend the method into a multi-stage progressive distillation to mitigate the loss of informative weights during extreme compression. Comprehensive experiments demonstrate the efficacy of TinyCLIP, showing that it can reduce the size of the pre-trained CLIP ViT-B/32 by 50\%, while maintaining comparable zero-shot performance. While aiming for comparable performance, distillation with weight inheritance can speed up the training by 1.4 - 7.8$\times$ compared to training from scratch. 
Moreover, our TinyCLIP ViT-8M/16, trained on YFCC-15M, achieves an impressive zero-shot top-1 accuracy of 41.1\% on ImageNet, surpassing the original CLIP ViT-B/16 by 3.5\% while utilizing only 8.9\% parameters. Finally, we demonstrate the good transferability of TinyCLIP in various downstream tasks. Code and models will be open-sourced at \href{https://aka.ms/tinyclip}{aka.ms/tinyclip}.

\end{abstract}

\vspace{-2mm}
\section{Introduction}

Large-scale language-image pretraining, \emph{e.g.}, 
 CLIP \cite{clip}, has recently gained significant attention due to its remarkable zero-shot transfer capability \cite{clip} and unprecedented performance in text-to-image generation \cite{DALLE-2}. Due to complex nature of vision and language, current approaches often resort to utilizing huge amounts of parameters to endow models with cross-modal capabilities \cite{clip, Align, coca, Florence, basic, LIMoE, flamingo}. This in turn leads to high costs in terms of storage, memory, and computation time for these models, which motivates the need for model compression to make them smaller and faster for real-world applications \cite{tinyvit,compressing}.

As a core compression technique, knowledge distillation has been extensively studied and applied in single-modal settings \cite{hilton_KD,kdsurvey}. However, its potential for multi-modality remains underexplored. Unlike single-modal models, the distillation of language-image cross-modal models poses distinct challenges. First, CLIP-like language-image models commonly consist of two branches: an image encoder and a text encoder  \cite{Align, Florence, flava, xvlm}. When distilling such multi-branch models, it is crucial to consider the interaction of information across the different modality branches in both the teacher and student models. Second, the original CLIP~\cite{clip} models are pre-trained on 400 million image-text pairs for 32 epochs, which requires thousands of GPU days, making distillation a significant challenge when computational resources are limited. Is there any way to reduce the cost of CLIP distillation? 
\begin{figure}[!t]
\vspace{-5mm}
\centerline{\includegraphics[width=0.47\textwidth]{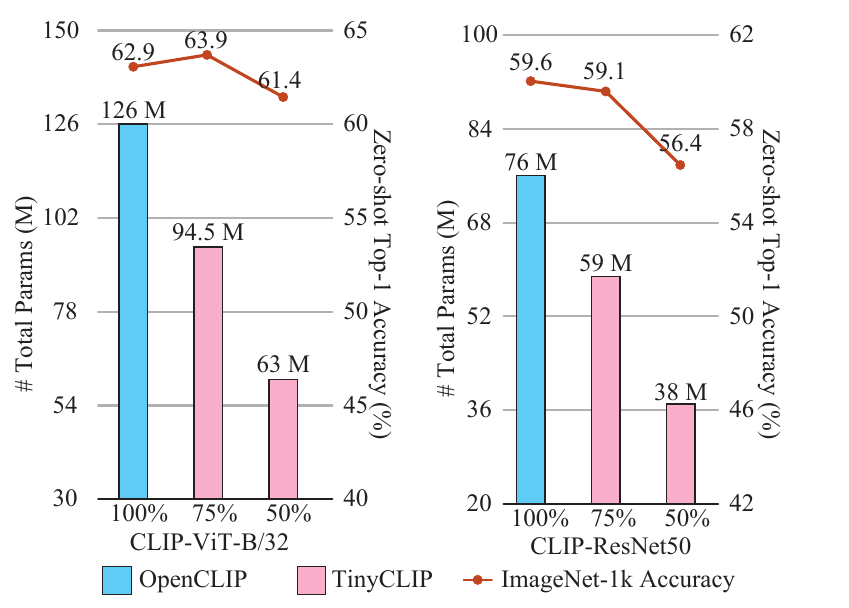}}
    \caption {Comparison of the OpenCLIP \cite{open_clip} and TinyCLIP. {TinyCLIP is pre-trained and distilled on LAION-400M  \cite{laion400m} using OpenCLIP ViT-B/32 \cite{open_clip} as the teacher, whose zero-shot top-1 accuracy is 65.6\%}. 
    }
    \label {fig:fig1}
    \vspace{-5mm}
\end{figure}

To tackle these challenges, we present a novel cross-modal distillation method dubbed TinyCLIP, which introduces two key techniques: \emph{affinity mimicking} and \emph{weight inheritance}. In contrast to the methods that rely on either image or text features for distillation, we empirically show that distilling knowledge in an image-text affinity space is more effective. Specifically, we leverage the cosine similarity of the image and text embeddings in the teacher model to facilitate the distillation of the student model, allowing the student to mimic the teacher's visual-linguistic feature alignment. We refer to this process as \emph{affinity mimicking}. 

To improve distillation efficiency, we introduce \emph{weight inheritance}, a technique transferring the pre-trained weights from teacher models to their student counterparts. Since inheriting weights provides a good initialization for the student models, the distillation progress can be largely accelerated. The key challenge of weight inheritance lies in determining which weights are more advantageous.
To address this issue, we introduce two solutions: manual and automatic inheritance. We surprisingly found that a simple manual selection of $k$-dimension or $k$-layer weights from the teacher model can yield satisfactory results for CLIP distillation. On the other hand, we also introduce learnable masks to automatically identify the most important weights from the teacher model. The masks are imposed independently on the vision and language branches, enabling them to capture the differences across modalities.

Moreover, we extend the proposed weight inheritance to a multi-stage progressive procedure, where each subsequent stage automatically inherits the important weights from the preceding stages. We observed that when the teacher model exhibits higher performance and shares a similar architecture with the student, weight inheritance can provide better results. This is because significant architecture differences may undermine the learned weights when transmitting from the teacher to the student. 
Therefore, we break the inheritance into multiple stages, allowing the student model in each stage to share a more similar structure with the predecessor teacher and inherit the weights progressively. 

Our experiments show that TinyCLIP delivers competitive models at all levels of speedups and model sizes in ImageNet zero-shot evaluation and various downstream tasks. As shown in Fig.~\ref{fig:fig1}, pre-trained on the same LAION-400M dataset  \cite{laion400m}, TinyCLIP-ViT using 63M parameters achieves 61.4\% zero-shot top-1 accuracy on ImageNet \cite{imagenet}, being 2.0$\times$ smaller and 1.5$\times$ faster than the OpenCLIP \cite{open_clip} model (62.9\% accuracy with 126M parameters). 
Meanwhile, TinyCLIP-ResNet with 38M parameters obtains 56.4\% zero-shot top-1 accuracy, being 2.1$\times$ smaller and 2.0$\times$ faster than CLIP ResNet-50 (59.6\% accuracy with 76M parameters). Moreover, our method can speed up the original OpenCLIP training by 1.4$\times$ -- 7.8$\times$ while achieving similar performance. Also, TinyCLIP demonstrates good transfer capacities in downstream scenarios. 
In summary, the contributions of this work are two-fold:
\begin{itemize}[leftmargin=0.468cm]
	\item{We propose a new cross-modal distillation approach to unleash the capacity of small CLIP models, fully leveraging large-scale models as well as pre-training data. To our best knowledge, this is the first work exploring the pre-training distillation of language-image  models.}
	
	\item{We present state-of-the-art language-image pre-trained models at small scale, striking the best trade-off between speed and accuracy. Extensive experiments demonstrate the superiority and good generalization ability of the small models in various downstream tasks.}
\end{itemize}

\section{Related Work}
\textbf{Language-Image Pre-training} has achieved remarkable progress over the past few years \cite{VLPsurvey, chen2023vlp}. In particular, contrastive language-image pre-training demonstrates very impressive zero-shot transfer and generalization capacities \cite{imagebert, clip, Align, Florence, simvlm, basic, lit}. One of the most representative works is CLIP \cite{clip}. A large amount of follow-up works have been proposed to optimize the pre-training framework \cite{ BLIP, regionclip, geimproving, unicl, icar, pyramidclip, ALBEF, tcl}. Meanwhile, a line of works leverage the pre-trained models for downstream tasks, such as open-vocabulary detection and segmentation \cite{glip, glipv2, simpleovseg,ovseg}, video recognition \cite{actionclip, ju2022prompting}, and text-to-image generation \cite{DALLE-2, styleclip}. 

Most recently, there are a few works attempting to scale language-image pre-trained models. BEiT-3 \cite{beit-3} uses a multiway transformer to scales up multi-modal pre-trained models to 1.9B parameters. FLIP \cite{flip} scales image-text pre-training via masked modeling, enabling CLIP to enjoy faster training speed while getting better performance. On the other hand, there are also few works on scaling down CLIP models \cite{upop}. However, they all focus on specific tasks, such as generation \cite{dai2022enabling, upop} and video content understanding \cite{clipping}. In contrast, our work concentrates on language-image pre-training, which is the first work on CLIP compression. 

\textbf{Knowledge Distillation} in a teacher-student framework \cite{hilton_KD} has been widely used to transfer knowledge from large models to small ones. Distillation in single modality, such as vision \cite{DeiT, tinyvit, teacher_assistant} or language \cite{distilbert, TinyBert}, has been extensively studied. However, the exploration of cross-modality distillation is relatively limited. A few early works exploring cross-modal distillation only focus on specific tasks, such as VQA and image caption \cite{dai2022enabling, fang2021compressing, cliptd}, limiting the methodologies to be applied in general downstream tasks. In contrast, this work pays attention to general cross-modal pre-training distillation, in which the distilled models can be transferred to various downstream scenarios.

It is worth noting that our proposed weight inheritance strategy shares similar spirits to model pruning techniques \cite{2022compressionsurvey,cofi, upop}. Both approaches involve the identification of important weights while removing redundant ones. However, they have three fundamental differences. 1) 
Previous pruning methods predominantly focus on single-modality models \cite{2022compressionsurvey, cofi, vit_slim}, whereas our approach shifts attention to multi-modality. Since different modalities exhibit varying redundancy characteristics, it is necessary to consider these differences when selecting important weights from each modality. 2) Unlike previous pruning methods which have emphasized the importance of the pruned architecture \cite{lottery, liurethinking, pruningfromscratch}, we argue that the weights inherited from the original models are also highly beneficial, particularly when compressing CLIP-like language-image models. 3) Our method leverages a progressive multi-stage process, which enables the inherited weights to accelerate the convergence of small models during training. 

\section{Method}
In this section, we propose TinyCLIP, a simple and effective method for distilling large-scale language-image pretraining models, such as CLIP.
It consists of three components: affinity mimicking (\S\ref{sec:Distillation}), weight inheritance (\S\ref{sec:Weightinheritance}) and multi-stage progressive distillation (\S\ref{sec:Multi-stage-compress}).

\subsection{Distillation with Affinity Mimicking}
\label{sec:Distillation}
A language-image pre-training model often consists of two branches, an image encoder to extract visual representations and a text encoder to extract textual representations.
The visual and textual representations are linearly projected into a cross-modal common space by minimizing a contrastive loss $\mathcal{L}_0$. 
The supervision signal can be viewed as an identity matrix, in which diagonal values are set to 1 for positive image-text pairs, and all others are set to 0 for negative pairs. This signal disregards the similarity between negative pairs, 
which prevents the model from acquiring a nuanced understanding of the intricate relationships among  negative pairs.
Therefore, we introduce \emph{affinity mimicking} to enable student models to learn similarities between negative pairs from teacher models. 

\begin{figure}[t]
\centerline{\includegraphics[width=0.42\textwidth]{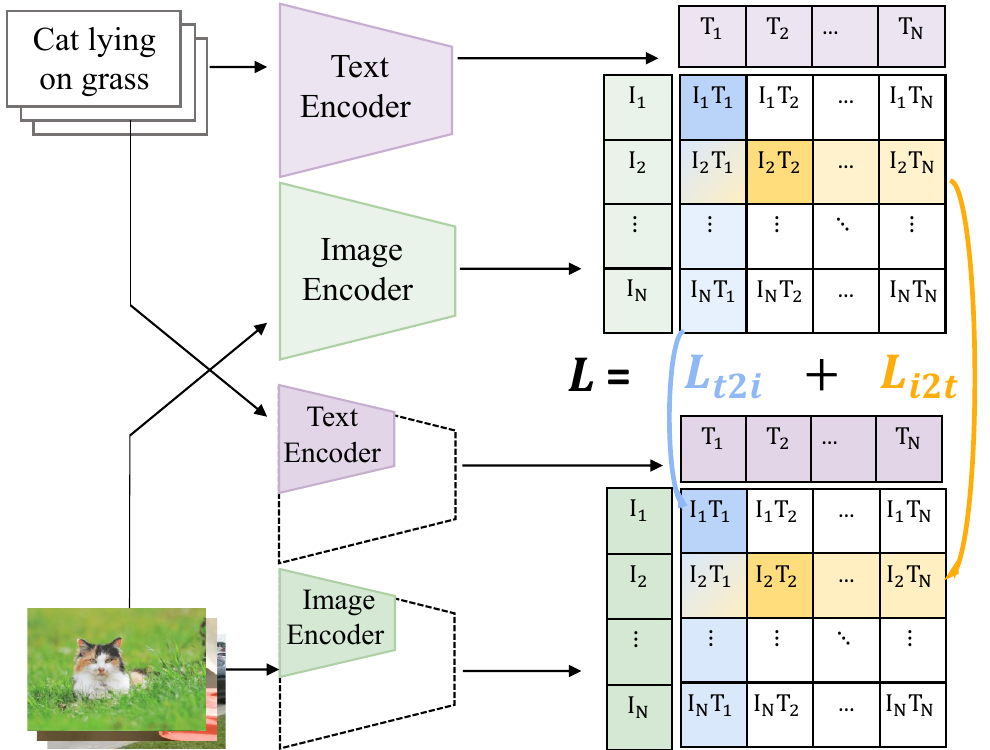}}
    \vspace{2mm}
    \caption {Affinity mimicking for language-image models. The loss includes image-to-text loss (yellow) and text-to-image loss (blue). }
    \label {fig:distillation}
\end{figure}

More specifically, as shown in Fig.~\ref{fig:distillation}, we consider two types of affinity distillation losses: image-to-language loss $\mathcal{L}_{I 2 T}$ and language-to-image loss $\mathcal{L}_{T 2 I}$. 
The former learns the alignment between the teacher and student models based on the image-to-language affinity $A_{I 2 T}$, which represents the affinity scores of an image with all the text descriptions in a batch (the yellow row highlighted in Fig.~\ref{fig:distillation}). The latter refers to the alignment of language-to-image affinity $A_{T 2 I}$, which compares a text description with all the images to obtain a matching score (the blue column in Fig.~\ref{fig:distillation}). The combination of these two losses formulates our affinity mimicking distillation, which is represented as: 
\begin{align}
\label{eq:distillation}
{\mathcal{L}}_{distill} &= \mathcal{L}_{I 2 T} + \mathcal{L}_{T 2 I}, \\
&= CE(A_{I 2 T}^s, A_{I 2 T}^t) + CE(A_{T 2 I}^s, A_{T 2 I}^t). \nonumber
\end{align}
Here, $CE$ represents the cross-entropy loss. The superscripts $s$ and $t$ denote the student and teacher models, respectively.
The elements in $A_{I 2 T}$ and $A_{T 2 I}$ are defined as: 
\begin{align}
A_{I 2 T}(i, j)&=\frac{\exp \left(I_i \cdot T_j / \tau\right)}{\sum_{k \in \mathcal{B}} \exp \left(I_{{ }_i}{ } \cdot T_k / \tau\right)},
\\
A_{T 2 I}(i, j)&=\frac{\exp \left(I_i \cdot T_j / \tau\right)}{\sum_{k \in \mathcal{B}} \exp \left(I_{{ }_k}{ } \cdot T{ }_j / \tau\right)},
\label{eq:sim}
\end{align}
where $\tau$ is a temperature parameter, $I_i$ is the feature embedding of $i$-th image in the batch $\mathcal{B}$, $T_j$ is the feature of $j$-th text description in $\mathcal{B}$. Affinity mimicking allows student models to imitate the behavior of the large model in learning visual-linguistic alignment. In contrast to previous methods that rely on either image or text
features for distillation, we show that distilling
knowledge in this image-text affinity space is more effective. 

\subsection{Distillation with Weight Inheritance}

\label{sec:Weightinheritance}

The original CLIP models are pre-trained on 400 million image-text pairs for 32 epochs, taking thousands of GPU days. This presents a significant cost challenge for distillation. To improve the training efficiency, we introduce weight inheritance, a technique that inherits the important weights from the well-trained large teacher models to smaller student models. 
The key challenge of inheriting weights is identifying important weights from the large amount of weights of the teacher.
We propose two approaches to select the important weights: manual weight inheritance and automatic weight inheritance. 

\textbf{Manual Inheritance}. 
For manual weight inheritance, we first analyze the redundancy of existing CLIP pre-trained models. Fig.~\ref{fig:feature_similar} shows that the text encoder displays more redundancy in depth (layer-wise), while the image encoder exhibits more redundancy in width (channel-wise). Based upon this discovery, we adopt the approach of uniformly selecting $k$ layers of the text branch and directly taking the front $k$ channels of the image branch to select the important weights from large teacher models. These weights will serve as the initialization for  small models during distillation. We surprisingly found that such a simple manual selected weights can largely accelerate CLIP distillation.

\begin{figure}[!t]
    \centerline{\includegraphics[width=0.45\textwidth]{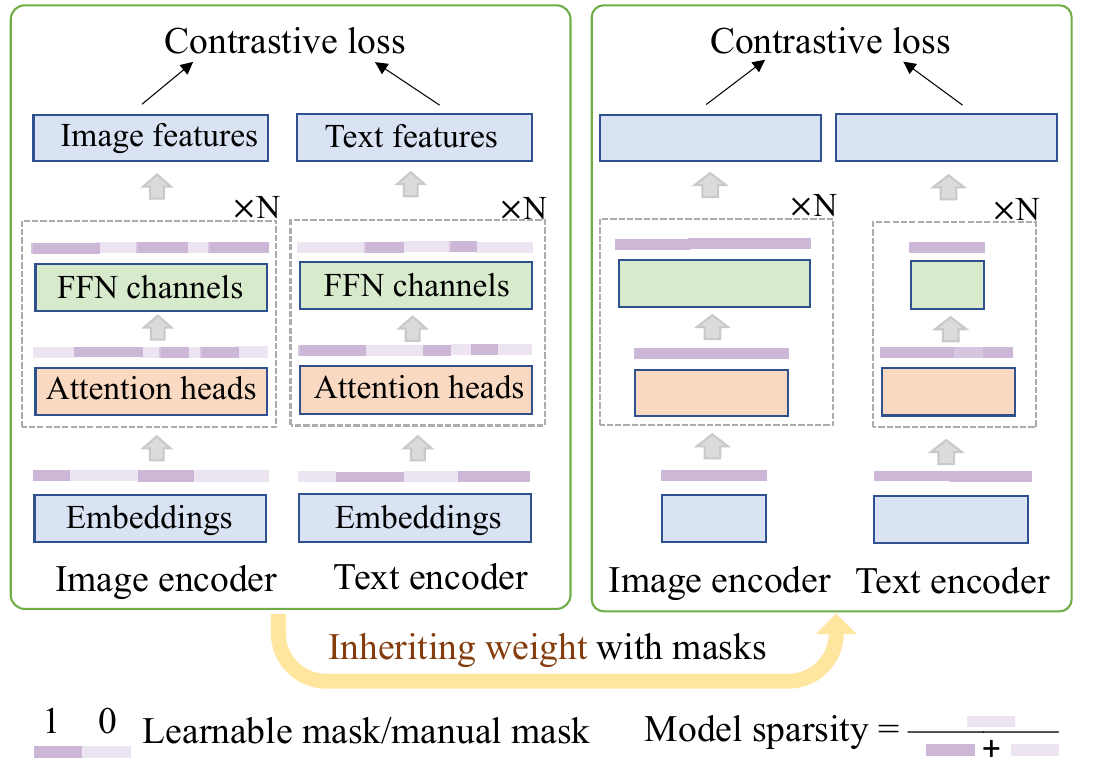}}
    \caption {Weight inheritance of CLIP. Given a large pre-trained CLIP model, the inherited weights are selected by masks. The weights whose mask is 0 will be removed.
    For the manual way, the masks are set to 1 in the first k-dimension or uniform k-layer, the rest is set to 0. For the automatic way, the masks are learned jointly with the model under the sparsity constraint.}
    \label {fig:weightinheritance}
    \vspace{-6mm}
\end{figure}

\textbf{Automatic Inheritance}.
Although manual inheritance can be considered an effective approach for model compression, it does present the drawback of necessitating prior knowledge in order to determine which weights to inherit. This limitation may render the technique impractical for application to diverse models.
To solve this issue, we present an automatic weight inheritance scheme. Inspired by structured pruning in large language models \cite{cofi},
we introduce learnable masks to identify weight importance.
Considering the difference across modalities, the learnable masks are imposed on vision and language branches independently, as visualized in Fig.~\ref{fig:weightinheritance}. An overall sparsity constraint is introduced to guarantee the selected  number of important weights to meet our compression requirements. 
Without loss of generality, here we use the transformer architecture as an example to introduce the detailed procedure of weight inheritance. Specifically, a standard transformer block contains a multi-head attention (MHA) layer and a feed-forward network (FFN). To capture the importance of weights in a fine-grained level, we introduce two mask variables ${\mathbf{m}}_\mathrm{head}$,  ${\mathbf{m}}_\mathrm{int}$ $\in$ $\{0,1\}$ to identify the redundant attention heads in MHA and  neurons in FFN respectively, while keeping the important ones. These two kinds of masks are imposed on the activation of attention heads and the intermediate layer of FFN, which is formulated as 
\begin{align}
    \mathrm{MHA}(\mathbf{X}) &= { \sum_{h=1}^{N_H}}  \mathbf{m}_{\mathrm{head}}^{h} \cdot \mathrm{Attn}_{\mathbf{W}_Q^{h}, \mathbf{W}_K^{h}, \mathbf{W}_V^{h}, \mathbf{W}_O^{h}}(\mathbf{X}),\\
    \mathrm{FFN}(\mathbf{X}) &=  \mathrm{GeLU}(\mathbf{X} \mathbf{W}_U) \cdot \mathrm{diag}(\mathbf{m}_{\mathrm{int}}) \cdot \mathbf{W}_D,
\end{align}
where $\mathbf{X}$ is the layer input, $\mathbf{W}_Q^{h}, \mathbf{W}_K^{h}, \mathbf{W}_V^{h}, \mathbf{W}_O^{h}$ $\in$ $ \mathbb{R}^{d \times d_h}$ denote the query, key, value and output matrices in MHA respectively, and $\mathbf{W}_U$$\in$$\mathbb{R}^{d \times d_f}$ and $\mathbf{W}_D$$\in$$\mathbb{R}^{d_f \times d}$ represent the parameters of FFN layers. Here, $d$ denotes the hidden dimension (\emph{e.g.}, 768), and $d_h$ = $d/N_H$ denotes the output dimension of each head (\emph{e.g.}, 64), where $N_H$ is the total number of heads.  
The trainable masks ${\mathbf{m}}_\mathrm{head}$ and ${\mathbf{m}}_\mathrm{int}$ serve as importance indicators for MHA and FFN layers. Moreover, to further learn the importance of embedding dimensions in transformer, we introduce an additional mask ${\mathbf{m}}_\mathrm{embed}$ $\in$ $\{0,1\}$. This mask is shared across all layers because each dimension in the hidden representation is connected to the corresponding dimension in the subsequent layer through a residual connection. 

During automatic inheritance, there are two losses 
used for optimizing
the masks and the model jointly, \emph{i.e.}, a sparsity loss and a distillation loss defined in Eq.~(\ref{eq:distillation}):
\begin{align}
\mathcal{L} &= \mathcal{L}_{distill} + \mathcal{L}_{sparsity},
\label{eq:loss_all}
\end{align}
\begin{align}
\mathcal{L}_{sparsity} &= \lambda \cdot (p - q) + \beta \cdot (p - q)^2,
\label{eq:loss_sparisty}
\end{align}
where $\lambda, \beta$ are learnable multipliers that guarantee  $p = q$\cite{wang2020structured,cofi}. Here, $q$ is the target compression rate, $p$ is the overall compression rate of learnable masks for the model, including image encoder and text encoder:
\begin{align}
&\scalebox{0.95}{$
p  = \frac{1}{M_{i}+M_{t}}\sum_{img, txt} ( 4 \cdot d_h \cdot \sum_i^L \sum_j^{N_H} \sum_k^d \mathbf{m}_{\text {head }}^{i, j} \cdot \mathbf{m}_{\text {embed }}^{k}$ \nonumber }
\\
&~~~\scalebox{0.95}{$
 + 2 \cdot \sum_i^L \sum_j^{d_f} \sum_k^d \mathbf{m}_{\mathrm{int}}^{i, j} \cdot \mathbf{m}_{\text {embed }}^{k}),$
}
\label{eq:sparsity}
\end{align}
where $M_i$ and $M_t$ represent the full model size of the image encoder and text encoder, respectively, $L$ is the total number of transformer layers. 

\subsection{Progressive Multi-Stage Distillation}
\label{sec:Multi-stage-compress}
When attempting to achieve a high target sparsity, \emph{i.e.}, \textgreater70\%, compressing the model in a single stage can lead to a significant reduction in accuracy and even result in convergence failure. This is due to the fact that most weights of the large model are directly discarded, including those that are important for ensuring model quality and convergence. 
As a solution, we propose a multi-stage progressive distillation method to achieve a high compression rate without seriously sacrificing accuracy. In each stage, we use a modest degree of compression, \emph{e.g.}, 25\%, to avoid large loss of performance and make training stable.

Specifically, each stage includes weight inheritance and affinity mimicking, as described in Algorithm~\ref{pseudocode:tinyclip}. During the weight inheritance phase, the weights of the large model are gradually reduced until the target sparsity and key weights are retained for the small model. After that, the small model is distilled with affinity mimicking, which involves transferring the visual-linguistic affinity knowledge from the teacher to the student. This procedure is repeated until the target sparsity level is achieved.

\begin{algorithm}[t]
\resizebox{0.93\linewidth}{!}{%
\caption{\small {TinyCLIP: Cross-Modal Distillation}}
\label{pseudocode:tinyclip}
\begin{threeparttable}
\footnotesize {
\LinesNumbered
\SetAlgoLined
\KwIn{a pre-trained CLIP model $f(x; \theta_0)$ with weight  $\theta_0$, target compression rate $q$, total iteration steps $N$}
Initialize the number of stages $G$, each stage steps $L$, target compression rate $q_i$ of stage $i$, weight inheritance steps $L_M$, input $x$, learnable mask $M=\{\mathbf{m}_\mathrm{head}, \mathbf{m}_\mathrm{int}, \mathbf{m}_\mathrm{embed}\}$;\tnote{1}

\For{$i$ in $[1,\dots ,G]$}{
// Increase mask compression rate, retain important weights\\
\For{$j$ in $[1,\dots ,L_M]$}{
Update target compression rate\tnote{2} $q$ of current step based on $q_i$ and step $j$; \\
Calculate \textbf{distillation loss} between origin teacher model $f(x; \theta_0)$ and masked student model $f(x; \theta_{i-1}\odot M_i)$;\\
Calculate \textbf{sparsity loss} between learnable mask compression rate\tnote{3} $p$ and target compression rate $q$;\\
Optimize $\theta_{i-1}$ and $M_i$;
}
$\theta_i \leftarrow \theta_{i-1} \odot M_i$ // Inherit weight with masks \\
// Cross-Modal distillation \\
\For{$j$ in $[L_M + 1,\dots , L]$}{
Calculate \textbf{distillation loss} between origin teacher model $f(x; \theta_0)$ and compressed model $f(x; \theta_i)$;\\
Optimize $\theta_i$;
}
}
\textbf{Return} compressed model $f(x; \theta_G)$ at compression rate $q$
}
\begin{tablenotes}
\footnotesize{
\item[1] {Each stage compress 25\%, $q_i = 1 - i \cdot 25\%$, $G = \lceil{(1-q)/0.25}\rceil$, $L =N/G$, $L_M =3000$, $M$ is initialized to $\textbf{1}$, $q_i$ can be adjusted according to different models.} \\
\item[2] {Target compression rate $q$ increases linearly from 0 to $q_i$ with step $j$.}
\item[3] {$p$ is calculated by masks $M$ in Eq.~(\ref{eq:sparsity}).}} \\
\end{tablenotes}
\end{threeparttable}
}
\end{algorithm}
\vspace{-1mm}

\section{Experiments}
\vspace{-1mm}
\subsection{Implementation Details}

\textit{Architecture.} The original CLIP employs a transformer-based model as its text encoder, while its image encoder has two variations, \emph{i.e.}, ViT  \cite{ViT} and modified ResNet~\cite{resnet}, covering both transformer-based and CNN-based architectures.
For weight inheritance, we inherited the weights from two model variants:  \emph{OpenCLIP} ViT-B/32 or ViT-B/16 \cite{open_clip} and \emph{CLIP} ResNet-50 \cite{clip}, which are pre-trained on LAION-2B \cite{laion5b} and WIT-400M \cite{clip}, respectively\footnote{The models can be downloaded from \href{https://github.com/mlfoundations/open_clip}{OpenCLIP} and \href{https://github.com/openai/CLIP}{OpenAI.}}.
For distillation, we use \emph{OpenCLIP} ViT-B/32  \cite{open_clip} pre-trained on LAION-2B \cite{laion5b} as the teacher model, since it achieves a high zero-shot accuracy of 65.6\% on ImageNet \cite{imagenet} with a high inference throughput. 
Besides, we also extend our method to DaViT~\cite{davit}, which is a hybrid architecture. We re-produce Florence-DaViT-5M \cite{Florence, davit} on LAION-400M as a baseline. Then we distill the same model using affinity mimicking and weight inheritance. The teacher model is Florence-DaViT-D3, which is pre-trained on FLD-900M \cite{Florence} and achieves a top-1 accuracy of 78.0\% on ImageNet \cite{imagenet} with 128M parameters.

\textit{Affinity Mimicking.} As defined in Tab.~\ref{tab:interaction}, we explore different interaction schemes between teacher and student models across modalities. The contrastive loss $\mathcal{L}_0$ and affinity mimicking $\mathcal{L}_1$ (\emph{i.e.}, $\mathcal{L}_{distill}$) serve as the basic ones, which can be combined with other losses. The weight of each loss is set to 1 when combined.
The temperature parameter $\tau$ in Eq.~(\ref{eq:sim}) is set to $\frac{1}{50}$ by default. The sensitivity analysis of $\tau$ is presented in the \emph{supplementary material}.

\textit{Weight Inheritance.} We compress OpenCLIP ViT-B \cite{open_clip} and CLIP ResNet-50 \cite{clip} using the proposed multi-stage progressive distillation. In each stage, we apply manual or automatic weight inheritance. \kanfix{
In manual inheritance, we compress the image encoder and the text encoder separately, by reducing the width of image encoder and the depth of text encoder. In automatic inheritance, we set the target sparsity for the whole model. In each  stage, the trainable masks are initialized to 1 and updated in the first 3,000 training iterations, where the optimizer is AdamW \cite{adamw, Adam} with a constant learning rate of 0.01 and no weight decay. The learnable multipliers $\lambda,\beta$ defined in Eq.~(\ref{eq:loss_sparisty}) are both initialized to 0.01.}

\textit{Training Settings.} We train our models on two public datasets, namely LAION-400M \cite{laion400m} and YFCC-15M \cite{yfcc100m}. On LAION-400M, the models are compressed in 3 stages, including 100\% to 75\% parameters for 6 epochs, 75\% to 50\% parameters for 16 epochs, and 50\% to 25\% parameters for 16 epochs. On YFCC-15M, it contains 2 compression stages, where the training epochs are both 25 from 100\% to 50\% parameters, and 50\% to 10\%.
\kanfix{We follow the hyper-parameter of CLIP \cite{clip} except that the learning rate is set to $10^{-4}$ when using weight inheritance. The details are shown in the \textit{supplementary material}.}
Unless otherwise specified, our ablation studies utilize the same settings, except for training the models only for 1 epoch with a learning rate of $5\times10^{-4}$.
All models are trained on 32 Nvidia V100 or A100, and implemented with PyTorch \cite{pytorch}, OpenCLIP  \cite{open_clip}, gradient cache \cite{grad_cache}, and Timm library \cite{timm}.

\textit{Evaluation Settings.} The models are evaluated on multiple benchmarks. In zero-shot transfer evaluation and robustness evaluation, we follow the same prompt engineering in CLIP \cite{clip}, where 80 text templates per class are used for ImageNet \cite{imagenet}. In linear probe, we use Elevater toolkit \cite{elevater} for evaluation, where a classification head is trained for 50 epochs with searched hyper-parameters.
Note that we do not count the number of parameters in the text embedding layer. It is a look-up table whose parameter size is the same as the models with the same hidden dimension and vocabulary size.  
The inference throughput is measured on Nvidia V100 with CUDA 11 and PyTorch v1.12 \cite{pytorch}, where the batch size is 1,024.

\begin{table*}[htbp]
\centering
\small
 \vspace{-1em}
\resizebox{0.99\textwidth}{!}{
\begin{tabular}{@{}lcccccccccccc@{}}
    \toprule
    \multirow{2}{*}{Method} & Image & \multicolumn{2}{c}{Text Encoder} & \#Params~(M) & MACs &  Throughput & Training & IN-1K & \multicolumn{2}{c}{Flickr30k} & \multicolumn{2}{c}{MSCOCO} \\
    \cmidrule(lr){3-4}
    \cmidrule(lr){10-11}
    \cmidrule(lr){12-13}
    ~ & Encoder & depth & width & Image+Text & (G) & (pairs/s) & datasets & top1 acc~(\%) & I $\rightarrow$ T@1 & T $\rightarrow$ I@1 & I $\rightarrow$ T@1 & T $\rightarrow$ I@1\\

    \midrule

    \multicolumn{5}{l}{\textit{Training data: 15M / 20M}} \\

    RILS \cite{rils} & ViT-B/16 & 12 & 512 & 86 + 38 & 20.5 & 818 & LAION-20M  & 45.0 & 45.1 & 34.9 & 32.2 & 25.5 \\  
    MaskCLIP \cite{maskclip} & ViT-B/16 & 12 & 512 & 86 + 38 & 20.5 & 818 & LAION-20M & 46.6 & 64.9 & 48.1 & 38.5 & 24.8 \\ 

    SLIP \cite{slip} & ViT-B/16 & 12 & 512 & 86 + 38 & 20.5 & 818 & YFCC-15M & 42.8 & 57.6 & 40.1 & 31.1 & 20.3 \\ 

    CLIP \cite{clip,slip} & ViT-B/16 & 12 & 512 & 86 + 38 & 20.5 & 818 & YFCC-15M & 37.6 & 51.6 & 32.2 & 26.5 & 17.1 \\ 
    
    \rowcolor{gray!10}
    TinyCLIP~(\textbf{Ours}) & ViT-39M/16 & 6 & 512 & 39+19(\textcolor{a_color}{2.1$\times$})  & 9.5(\textcolor{a_color}{2.2$\times$}) & 1,469(\textcolor{a_color}{1.8$\times$}) & YFCC-15M & 63.5 & 84.4 & 66.7 & 54.9 & 38.9 \\ 

    \rowcolor{gray!10}
    TinyCLIP~(\textbf{Ours}) & ViT-8M/16 & 3 & 256 & 8+3(\textcolor{a_color}{11.3$\times$})  & 2.0(\textcolor{a_color}{10.3$\times$}) & 4,150(\textcolor{a_color}{5.1$\times$}) & YFCC-15M & 41.1 & 62.3 & 42.3 & 36.2 & 21.5 \\ 
    
    \midrule

    \multicolumn{5}{l}{\textit{Training data: 400M}}\\
    
    Florence \cite{Florence} & DaViT-5M & 6 & 256 & 5+5 & 1.1 & 2,980 & LAION-400M & 45.0 & 61.2 & 41.5 & 36.2 & 20.9 \\
    \rowcolor{gray!10}
    TinyCLIP~(\textbf{Ours}) & DaViT-5M & 6 & 256 & 5+5 & 1.1 & 2,980 & LAION-400M & 50.0 & 66.2 & 48.5 & 40.7 & 24.5 \\
    
    \hdashline

    CLIP \cite{clip} & ResNet-101 & 12 & 512& 56 + 38 & 12.8 & 1,161 & WIT-400M & 62.2 & 78.1 & 59.2 & 49.3 & 30.7 \\
    
    CLIP \cite{clip} & ResNet-50 & 12 & 512& 38 + 38 & 9.1 & 1,549 & WIT-400M & 59.6 & 81.2 & 58.2 & 50.8 & 28.3 \\

    \rowcolor{gray!10}
    TinyCLIP~(\textbf{Ours}) & ResNet-30M & 9 & 512 & 30+29(\textcolor{a_color}{1.3$\times$}) & 6.9(\textcolor{a_color}{1.3$\times$}) & 1,811(\textcolor{a_color}{1.2$\times$}) & LAION-400M & 59.1 &
 81.1 & 61.2 & 52.7 & 33.9 \\    
    \rowcolor{gray!10}
    TinyCLIP~(\textbf{Ours}) & ResNet-19M & 6 & 512 & 19+19(\textcolor{a_color}{2.0$\times$}) & 4.4(\textcolor{a_color}{2.1$\times$}) & 3,024(\textcolor{a_color}{2.0$\times$}) & LAION-400M & 56.4 & 76.2 & 58.3 & 48.9 & 30.9 \\ 

    \hdashline
    OpenCLIP \cite{open_clip} & ViT-B/32 & 12 & 512 & 88 + 38 & 7.4 & 2,452 & LAION-2B & 65.7 & 84.7 & 66.8 & 56.9 & 39.3 \\
    CLIP \cite{clip} & ViT-B/32 & 12 & 512 & 88 + 38 & 7.4 & 2,452 & WIT-400M & 63.2 & 80.1 & 59.8 & 51.2 & 30.6 \\
    
    OpenCLIP \cite{open_clip} & ViT-B/32 & 12 & 512 & 88 + 38 & 7.4 & 2,452 & LAION-400M & 62.9 & 79.3 & 62.0 & 53.3 & 35.4\\

    \rowcolor{gray!10}
    TinyCLIP~(\textbf{Ours}) & ViT-61M/32 & 9 & 512 & 61+29(\textcolor{a_color}{1.4$\times$}) & 5.3(\textcolor{a_color}{1.4$\times$}) & 3,191(\textcolor{a_color}{1.3$\times$}) & LAION-400M & 62.1 & 78.6 & 63.3 & 53.9 & 35.9 \\ 
    
    \rowcolor{gray!10}
    TinyCLIP~(\textbf{Ours}) & ViT-40M/32 & 6 & 512 & 40+19(\textcolor{a_color}{2.1$\times$}) & 3.5(\textcolor{a_color}{2.1$\times$}) & 4,641(\textcolor{a_color}{1.9$\times$}) & LAION-400M & 59.7 & 77.3 & 58.9 & 49.8 & 33.1 \\ 

    \hdashline[1.1pt/6pt]
    \rowcolor{gray!10}
    TinyCLIP~(\textbf{Ours}) & ViT-63M/32 & auto & auto & 63+31(\textcolor{a_color}{1.3$\times$}) & 5.6(\textcolor{a_color}{1.3$\times$}) & 2,905(\textcolor{a_color}{1.2$\times$}) & LAION-400M & 63.9 & 83.2 & 64.4 & 55.5 &  37.6 \\ 
    
    \rowcolor{gray!10}
    TinyCLIP~(\textbf{Ours}) & ViT-45M/32 & auto & auto & 45+18(\textcolor{a_color}{2.0$\times$}) & 3.7(\textcolor{a_color}{2.0$\times$}) & 3,682(\textcolor{a_color}{1.5$\times$}) & LAION-400M & 61.4 & 80.9 & 62.2 & 52.8 & 34.7  \\
    \rowcolor{gray!10}
    TinyCLIP~(\textbf{Ours}) & ViT-22M/32 & auto & auto & 22+10(\textcolor{a_color}{3.9$\times$}) & 1.9(\textcolor{a_color}{3.9$\times$}) & 5,504(\textcolor{a_color}{2.2$\times$}) & LAION-400M & 53.7 & 71.3 & 52.0 & 44.4 & 28.3 \\
    \hdashline[1.1pt/6pt]

    \rowcolor{gray!10}
    TinyCLIP~(\textbf{Ours}) & ViT-63M/32 & auto & auto & 63+31(\textcolor{a_color}{1.3$\times$}) & 5.6(\textcolor{a_color}{1.3$\times$}) & 2,909(\textcolor{a_color}{1.2$\times$}) & LAION+YFCC-400M & 64.5 & 84.9 & 66.0 & 56.9 & 38.5 \\ %
    \rowcolor{gray!10}
    TinyCLIP~(\textbf{Ours}) & ViT-45M/32 & auto & auto & 45+18(\textcolor{a_color}{2.0$\times$}) & 3.7(\textcolor{a_color}{2.0$\times$}) & 3,685(\textcolor{a_color}{1.5$\times$}) & LAION+YFCC-400M & 62.7 & 80.3 & 63.9 & 54.0 & 36.7 \\ %

    \bottomrule
\end{tabular}
}
\vspace{2mm}
\caption{Comparison with the state-of-the-art methods. The architecture of text encoder is Transformer \cite{attention}. ``auto'' denotes automatic weight inheritance, and our other models use manual weight inheritance.}
\label{tab:main_table}
\end{table*}

\subsection{Comparison with State-of-the-Art Models}

We compare our TinyCLIP with state-of-the-art models in Tab. \ref{tab:main_table}. The zero-shot evaluation on ImageNet-1K \cite{imagenet}, Flickr30k \cite{plummer2015flickr30k} and MSCOCO retrieval \cite{coco} are reported.

When performing training on YFCC-15M \cite{yfcc100m}, we use affinity mimicking and manual weight inheritance to distill OpenCLIP ViT-B/16 \cite{open_clip}, whose zero-shot performance on ImageNet is 70.2\%. By inheriting half of parameters, our TinyCLIP ViT-39M/16 can achieve a zero-shot accuracy of 63.5\% on ImageNet, surpassing the original CLIP ViT-B/16 \cite{clip} by 25.9\%. 
Furthermore, when compressing the image encoder to 10 layers with 256 dimensions and the text encoder to 3 layers with 256 dimensions, TinyCLIP ViT-8M/16 still surpasses CLIP ViT-B/16 \cite{clip} by 3.5\%  while using  $11.3\times$ fewer parameters and running $5.1\times$ faster. 

When conducting training on LAION-400M \cite{laion400m}, we distill three different models, \emph{i.e.}, Florence DaViT \cite{Florence, davit}, CLIP ResNet \cite{clip}, and OpenCLIP ViT-B/32 \cite{open_clip}. The results in Tab.~\ref{tab:main_table} demonstrate the efficacy of our method. More specifically, our TinyCLIP DaViT-5M achieves up to 5.0\% improvements over the original model on ImageNet, and 5.0/4.5\% gains in image-to-text retrieval on Flicker and MSCOCO, respectively. It is worth noting that this is a tiny language-image pre-trained model using only 10M parameters to achieve 50.0\% accuracy on ImageNet.

For CLIP ResNet, TinyCLIP-30M is only trained for 4 epochs in the first stage, inheriting 75\% parameters from the original model. 
The results indicate that there is 0.5\% slight decrease in top-1 accuracy on ImageNet, however, has $\sim$2\% improvements in retrieval tasks. TinyCLIP ResNet-19M inherits the weights from the 30M model, and it is trained for 12 epochs. It reduces the parameters by 50\% while getting $2\times$ inference speedup. But it gets 2.7\% performance drops on ImageNet due to the large reduction in parameters.

For OpenCLIP ViT-B/32, we distill it using our affinity mimicking method with manual or automatic weight inheritance. We design a three-stage progressive distillation in which each stage compresses the model by 25\%. The three stages are trained for 6, 16, and 16 epochs, respectively.
In manual inheritance, TinyCLIP ViT-61M/32 obtains comparable performance compared to OpenCLIP ViT-B/32 \cite{open_clip}, while reducing the parameters by 29\%. On the other hand, automatic inheritance obtains better results than the manual one. For example, TinyCLIP ViT-63M/32 outperforms the manually inherited ViT-61M/32 model by more than $\sim$1\% on both classification and retrieval tasks. When auto-compressed the model by $3.9 \times$, TinyCLIP ViT-22M/32 obtains the highest inference throughput of 5,504 and reaches 53.7\% top-1 accuracy. Since YFCC-15M contains high-quality image-text pairs, if using it to replace parts of LAION data, our method can obtain additional improvements, especially for retrieval tasks.

\subsection{Ablation Study}
\label{ablation}

\textit{Impact of affinity mmicking.} As shown in Tab. ~\ref{tab:interaction}, we present four different interaction modes. The affinity mimicking $\mathcal{L}_1$ (\emph{i.e.}, $\mathcal{L}_{distill}$) outperforms the contrastive loss $\mathcal{L}_0$ by 2.1\% in terms of top-1 accuracy on ImageNet \cite{imagenet}. 
The alignment in a visual-linguistic affinity space brings more similarity information.
The accuracy of cross modalities interaction is lower than that of affinity mimicking by 0.2\%. In this mode, the embedding features of student image and text do not have any interaction, but they are aligned to the teacher's embedding space.
Single modality interaction has lower accuracy due to missing a loss to align student's image and text features in common embedding space.

\begin{table}[t]
\centering
\resizebox{1.0\linewidth}{!}{
\begin{tabular}{llc}
\toprule
Interaction Mode & Loss Formula & Top-1 Acc\\
\hline
Contrastive loss \cite{clip} & $\mathcal{L}_0 = CE(<I_s, T_s> , \mathcal{I})$ & 53.4  \%\\

\rowcolor{gray!10}
Affinity mimicking & $\mathcal{L}_1 = CE(<I_s, T_s>, <I_t, T_t>)$ & 55.5 \%\\

\multirow{2}{*}{Cross modalities} & $\mathcal{L}_2 = CE(<I_s, T_t>, <I_t, T_t>)$ & \multirow{2}{*}{55.3 \%} \\
~ & $\mathcal{L}_3 = CE(<I_t, T_s>, <I_t, T_t>)$ & ~ \\

\rowcolor{gray!10}
 ~ & $\mathcal{L}_4 = CE(<I_s, I_t>, \mathcal{I})$ & ~ \\
\rowcolor{gray!10}
\multirow{-2}{*}{\cellcolor{gray!10}Single modality} & $\mathcal{L}_5 = CE(<T_s, T_t>, \mathcal{I})$ & \multirow{-2}{*}{19.2 \%} \\

\bottomrule
\end{tabular}
}
\vspace{2mm}
\caption {The interaction of information across image $I$ and text $T$ modalities in student and teacher models. The subscripts $s$ and $t$ denote student and teacher, respectively. $\mathcal{I}$ is an identity matrix. $CE$ is cross entropy function. TinyCLIP-ViT-40M/32 is trained with an interaction mode and manual weight inheritance on LAION-400M \cite{laion400m} for 1 epoch. The zero-shot accuracy on ImageNet \cite{imagenet} is reported.}

\label{tab:interaction}
\end{table}


\begin{figure}[t]
        \resizebox{1.0\linewidth}{!}{
        \includegraphics[]{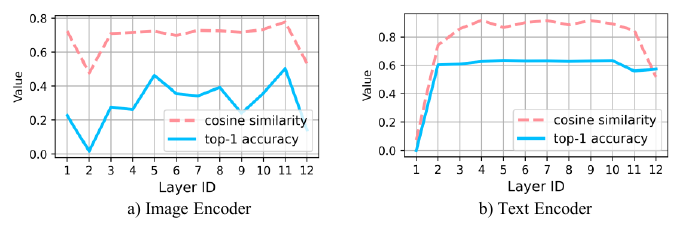}
        }
    \caption {The cosine similarity of the input and output embedding of each layer, and the zero-shot accuracy on ImageNet \cite{imagenet} after removing a layer for OpenCLIP ViT-B-32 \cite{open_clip, clip} pretrained on LAION-2B \cite{laion5b}.}
    \label {fig:feature_similar}
\end{figure}


\begin{table}[t]
\resizebox{1.01\linewidth}{!}{%

\begin{tabular}{cccccc}
\toprule
\multirow{2}{*}{Weight Inheritance} &
\multicolumn{1}{c}{IN-1K} & \multicolumn{2}{c}{Flickr30k} & \multicolumn{2}{c}{MSCOCO} \\
\cmidrule(lr){3-4}
\cmidrule(lr){5-6}
~ & top1 acc~(\%)  & I$\rightarrow$T@1 & T$\rightarrow$I@1 & I$\rightarrow$T@1 & T$\rightarrow$I@1 \\
\midrule
None & 37.8 & 50.2 & 33.4 & 29.7 & 16.6 \\ 
\hline
Manual & 54.0 & 70.4 & 52.9 & 46.5 & 29.3 \\ 
Automatic & 54.9 & 73.5 & 55.9 & 47.9 & 30.8 \\ 
\bottomrule
\end{tabular}
}
\caption{Manual inheritance \emph{vs.} Automatic inheritance. The model OpenCLIP ViT-B/32 \cite{open_clip} is compressed by 50\% with distillation on LAION-400M \cite{laion400m} for 1 epoch.}
\vspace{-2mm}
\label{tab:weight_inherit_methods}
\end{table}

\textit{Impact of weight inheritance.}
We verify the efficacy of the proposed two kinds of weight inheritance in Tab.~\ref{tab:weight_inherit_methods}. Compared to the model without using weight inheritance, manual inheritance brings $16.2\%$ accuracy improvements on ImageNet \cite{imagenet}. Automatic weight inheritance can further improve the accuracy by 0.9\%. On image-text retrievals, the improvements of automatic inheritance is relatively higher, \emph{i.e.}, 1.4 -- 3.1\% over the manual method.

\textit{The redundancy of language-image models.} We analyze the redundancy of image encoder and text encoder by similar strategy \cite{universal_transformer, Deep_equilibrium_models, Albert} and zero-shot accuracy by removing a layer. As shown in Fig.~\ref{fig:feature_similar}, we observe a high correlation between cosine similarity and accuracy. Comparing with image encoder, text encoder has higher cosine similarity. It shows the redundancy of the text encoder is higher than that of the image encoder. Consequently, compressing the text encoder along the depth dimension can be achieved without significant performance degradation.

\begin{figure}[t]
    \centering
    \centerline{\includegraphics[width=8cm]{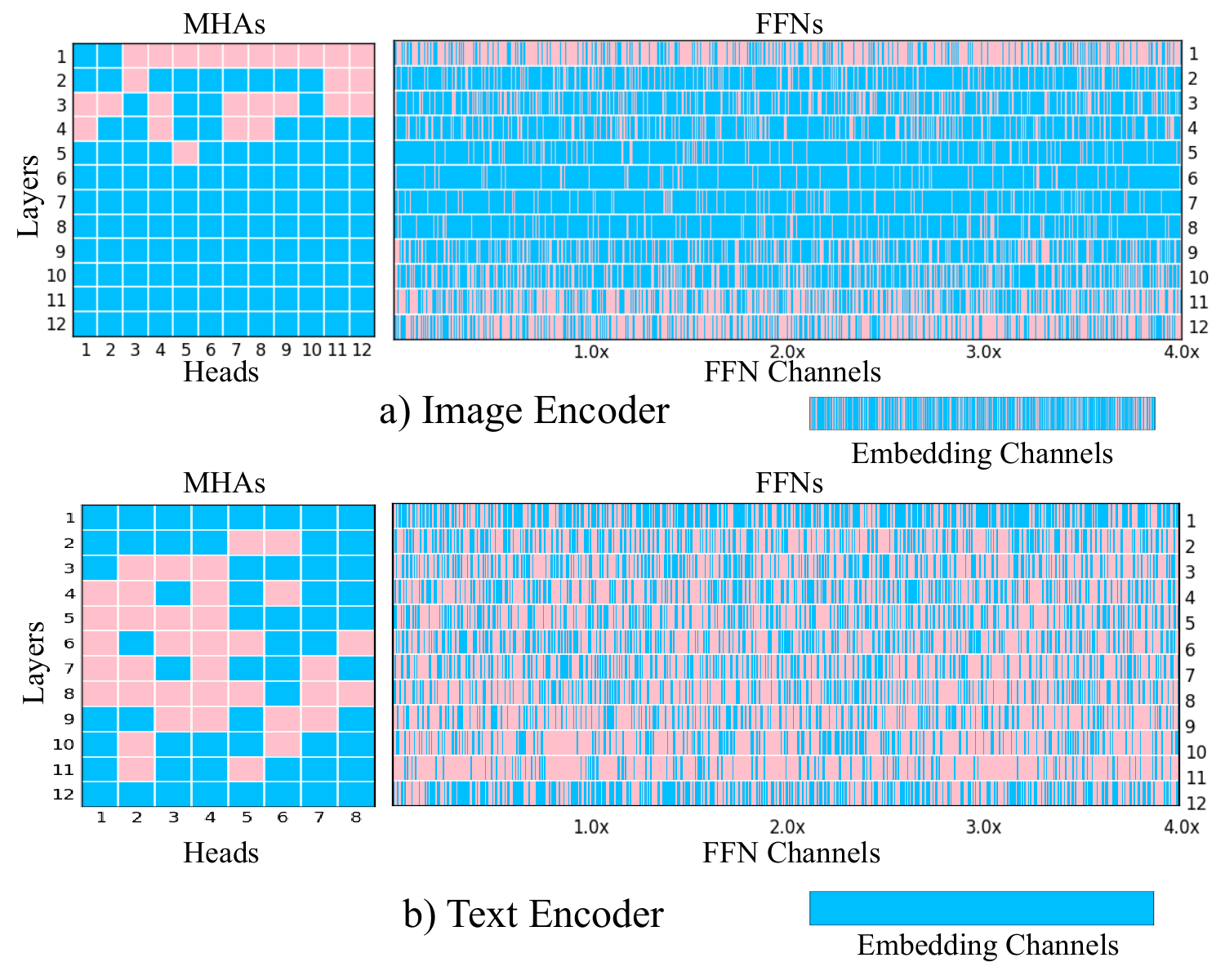}}
    \caption {Visualization of learnable masks at 50\% sparsity
    on hidden dimension, attention heads and MLP intermediate dimension. The blocks in red color are removed. The original model is OpenCLIP ViT-B/32 \cite{open_clip}.}
    \label {fig:maskvisual}
\end{figure}

We further explore the redundancy by the proposed automatic weight inheritance.
As shown in Fig.~\ref{fig:maskvisual},
for width analysis, it is observed that the embedding channels of the text encoder experience only a marginal reduction (\emph{e.g.}, 512 $\rightarrow$ 511), whereas the embedding channels of the image encoder are reduced to a greater extent (e.g., 768 $\rightarrow$ 526). 
Regarding depth analysis, it was observed that a significant proportion of MHA layers and FFN channels could be removed from the text encoder, indicating redundancy within it. This finding suggests that the image encoder can be compressed along the width dimension, while the text encoder can be compressed along the depth dimension.

\textit{Impact of teacher models.} We investigate which model is better for weight inheritance. As analyzed in Tab. \ref{tab:inherited_weights}, a strong teacher does not necessarily guarantee better results for weight inheritance. Instead, teacher models with similar architectures and higher performance offer a better choice.
For instance, although ViT-H/14 ranks as the highest-performing teacher model, it lags behind other models in terms of weight inheritance.

\begin{table}[t]
\Huge
\centering
\resizebox{0.8\linewidth}{!}{%
\begin{tabular}{lccc}
\toprule
Pretrained & Teacher~~~ & Inherited & ~~~~~~Student~~~~~~~
\\
Teacher Model & Acc.(\%) & Ratio & ~~~~~~Acc.(\%)~~~~~~ \\
\midrule
w/o weight inheritance & - & 0 & 36.2 \\  
CLIP ViT-B/32 \cite{clip} & 63.2 & 59 / 126 & 52.4(\textcolor{a_color}{+16.2}) \\
OpenCLIP ViT-B/32 \cite{open_clip} & 62.9 & 59 / 126  & 53.5(\textcolor{a_color}{+17.3}) \\
OpenCLIP ViT-B/16 \cite{open_clip}  & 67.1 & 59 / 124 & 52.8(\textcolor{a_color}{+16.6}) \\
OpenCLIP ViT-L/14 \cite{open_clip}  & 75.3 & 59 / 390   & 45.1(\textcolor{a_color}{+8.9}) \\
OpenCLIP ViT-H/14 \cite{open_clip}  & 78.0 & 59 / 935   & 41.1(\textcolor{a_color}{+4.9}) \\

\bottomrule
\end{tabular}
}
\vspace{2mm}
\caption{Ablation study on inherited teacher model. The student model TinyCLIP ViT-40M/32 with 59M parameters is inherited from a teacher model, then trained without distillation on LAION-400M \cite{laion400m} for 1 epoch. The top-1 zero-shot accuracy on ImageNet \cite{imagenet} is reported.}
\label{tab:inherited_weights}
\end{table}

\begin{table*}[t]
\vspace{-1em}
    \centering
    \setlength{\tabcolsep}{1pt}{
    \resizebox{1.0\textwidth}{!}{
    \begin{tabular}{cc|ccccccccccccccccccccccccccc}
\hline
        
        Method & \rotatebox[origin=l]{90}{\small Image Encoder}
        & \rotatebox[origin=l]{90}{\footnotesize Food101}
        & \rotatebox[origin=l]{90}{\footnotesize CIFAR10}
        & \rotatebox[origin=l]{90}{\footnotesize CIFAR100}
        & \rotatebox[origin=l]{90}{\footnotesize SUN397}
        & \rotatebox[origin=l]{90}{\footnotesize Stanford Cars}
        & \rotatebox[origin=l]{90}{\footnotesize FGVC Aircraft}
        & \rotatebox[origin=l]{90}{\footnotesize VOC2007}
        & \rotatebox[origin=l]{90}{\footnotesize DTD}
        & \rotatebox[origin=l]{90}{\footnotesize Oxford Pets}
        & \rotatebox[origin=l]{90}{\footnotesize Caltech101}
        & \rotatebox[origin=l]{90}{\footnotesize Flowers102}
        & \rotatebox[origin=l]{90}{\footnotesize MNIST}
        & \rotatebox[origin=l]{90}{\footnotesize FER2013}
        & \rotatebox[origin=l]{90}{\footnotesize STL10}
        & \rotatebox[origin=l]{90}{\footnotesize EuroSAT}
        & \rotatebox[origin=l]{90}{\footnotesize RESISC45}
        & \rotatebox[origin=l]{90}{\footnotesize GTSRB}
        & \rotatebox[origin=l]{90}{\footnotesize KITTI}
        & \rotatebox[origin=l]{90}{\footnotesize Country211}
        & \rotatebox[origin=l]{90}{\footnotesize PCam}
        & \rotatebox[origin=l]{90}{\footnotesize HatefulMemes}
        & \rotatebox[origin=l]{90}{\footnotesize Rendered SST2}
        & \rotatebox[origin=l]{90}{\footnotesize ImageNet} \\
\hline

    \multicolumn{1}{l}{\textit{Zero-shot performance}} &  &  &  &  &  &  &  &  &  &  &  &  &  &  &  &  &  &  &  &  &  &  &  & &  &  &  &  \\ 
    DeCLIP \cite{declip} & ViT-B/32 & 76.3 & 90.0 & 67.9 
    & 63.5 & 50.6 & 8.9 & 80.5 & 45.3 & 84.4 & 89.3 & 83.2 & 12.8 & 16.3 & 97.5 & 40.0 & 48.4 & 10.3 & 35.3 & 11.5 & 54.7 
    & 52.9 & 50.1 & 66.1 \\
    
    FILIP \cite{filip} & ViT-B/32 & 82.8 & 86.9 & 65.5
    & 69.1 & 55.5 & 57.2 & - & 49.3 & 88.1 & 91.9 & 85.3 & - & - & - & - & 49.9 & - & -  & - & - & - & - & 68.8 \\

    CLIP \cite{clip} & ViT-B/32& 84.4 & 91.3 &65.1 
    &63.2& 59.4& 21.2& 83.1& 44.5& 87.0& 87.9& 66.7& 51.9& 47.3& 97.2& 49.4& 60.3& 32.2& 39.4& 17.8& 58.4
    & 57.6& 59.6& 63.2 \\
    
    OpenCLIP \cite{open_clip} & ViT-B/32 & 80.9 & 90.7 & 70.6 & 66.8 & 79.3 & 16.6 & 82.2 & 54.4 & 86.5 & 90.1 & 66.0 & 37.4 & 42.3 & 95.6 & 51.6 & 57.6 & 42.0 & 31.6 & 14.8 & 50.1 & 52.9 & 52.3 & 62.9 \\
    
    \rowcolor{gray!10}
    \textbf{TinyCLIP (Ours)} & ViT-45M/32 & 78.3 & \underline{92.2} & \underline{71.8} & 66.0 & \underline{79.9} & 23.0 & \underline{83.2} & 53.8 & 85.3 & 88.7 & 60.7 & \underline{62.5} & 45.2 & 96.2 & \underline{52.2} & \underline{61.8} & 41.1 & 16.9 & 16.5 & 51.5 &  53.5 & 52.6 & 61.2  \\ 

    \hline
    \multicolumn{2}{l|}{\textit{Linear probe performance}} &  &  &  &   &  &  &  &  &  &  &  &  &  &  &  &  &  &  &  &    &  &  &  \\
    CLIP \cite{clip} & ViT-B/32 & 88.8& 95.1& 80.5
    & 76.6& 81.8& 52.0& 87.7& 76.5& 90.0& 93.0& 96.9& 99.0& 69.2& 98.3& 97.0& 90.5& 85.3& 66.2& 27.8& 83.9
    & 66.7& 70.8& 76.1 \\
    
    \rowcolor{gray!10}
    \textbf{TinyCLIP (Ours)} & ViT-45M/32 & 84.0 & \underline{97.4} & \underline{86.0} 
    & 68.9 & \underline{85.4} & 39.0 & 85.4 & 71.2 & 89.9 & \underline{93.4} & 82.1 & \underline{99.5} & 66.9 & 97.6 & \underline{97.5} & 88.1 & \underline{98.5} & \underline{79.7} & 16.1 & \underline{90.4} 
    & 58.4 & 62.5 & 73.2  \\
    
\bottomrule
        
    \end{tabular}
    }
    }
    \vspace{1mm}
    \caption{Zero-shot and linear-probe classification top-1 accuracy on 23 datasets.}
    \label{tab:zs-ic}
    \vspace{-2mm}
\end{table*}

\begin{table}[t]
\centering
\resizebox{1.0\linewidth}{!}{%

\begin{tabular}{cccccc}
\toprule
\multirow{1}{*}{Number} & \multirow{2}{*}{Model Size} & \multirow{2}{*}{Epoch} &
\multicolumn{1}{c}{ImageNet} & \multicolumn{2}{c}{MSCOCO} 
\\
\cmidrule(lr){5-6}
of Stages & ~ & ~ & top1 acc~(\%) & I$\rightarrow$T R@1& T$\rightarrow$I R@1\\
\midrule
1 & $100\% \rightarrow 25\%$ & 2 & 47.0 & 39.4 & 23.1\\
\midrule
\multirow{2}{*}{2} & $100\% \rightarrow 50\%$ & 1 & 54.0 & 46.5 & 29.3 \\
~ & $50\% \rightarrow 25\%$ & 1 & 49.0 & 40.2 & 25.3 \\
\bottomrule
\end{tabular}
}
\vspace{-1mm}
\caption{Ablation study on multi-stage progressive distillation. The model OpenCLIP ViT-B/32 \cite{open_clip} is compressed to 25\% model size on LAION-400M \cite{laion400m} for 2 epochs.}
\label{tab:multi_stage}
\vspace{-3mm}
\end{table}

\newcommand{\disgmask}{\Vert \frac{\partial{\mathcal{L}_{distill}}}{\partial{mask}} \Vert}
\newcommand{\spargmask}{\Vert \frac{\partial{\mathcal{L}_{sparsity}}}{\partial{mask}} \Vert}

\textit{Impact of multi-stage progressive distillation.}
Multi-stage progressive distillation shrinks the gap between the compressed model and the inherited model in each stage. We compare it with single-stage in Tab.~\ref{tab:multi_stage}. Using the same training epochs, the model trained with 2-stage surpasses that with 1-stage by 2.0$\%$ in terms of zero-shot accuracy on ImageNet. It demonstrates that multi-stage progressive compression is superior when transferring the knowledge of the pre-trained model to the compressed small model.

\begin{figure}[t]
  \centering
  \includegraphics[width=0.85\linewidth]{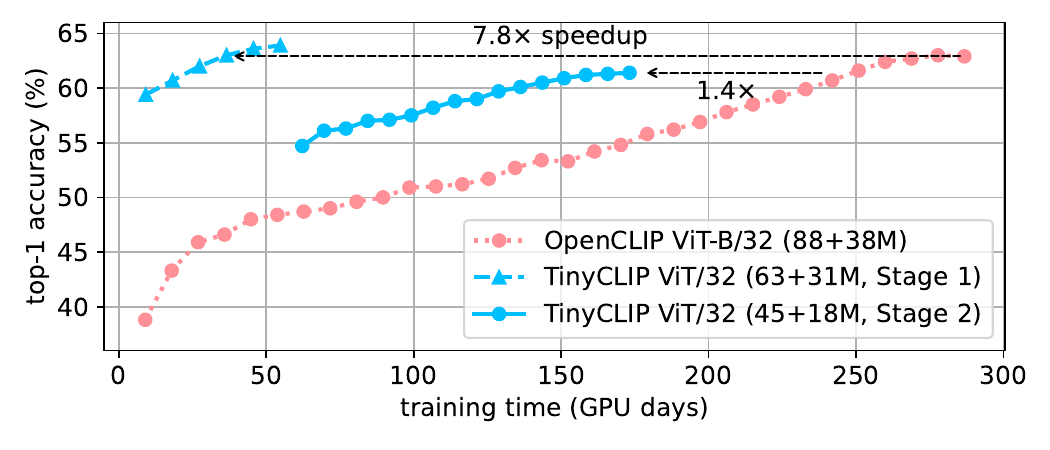}
  \caption{Training efficiency. OpenCLIP ViT-B/32 \cite{open_clip} is trained from scratch on LAION-400M \cite{laion400m} for 32 epochs. TinyCLIP ViT/32 is trained for 6+16 epochs using the proposed 2-stage progressive distillation, where the teacher is OpenCLIP ViT-B/32.} 
  \label{fig:acc_train_time}
  \vspace{-3mm}
\end{figure}

\textit{Training cost.} We also study the distillation efficiency. As shown in Fig. \ref{fig:acc_train_time}, the original OpenCLIP ViT-B/32 \cite{open_clip} takes around 287 GPU days when trained on LAION-400M for 32 epochs. Our models with 75\% and 50\% parameters reach the similar accuracy as OpenCLIP counterparts, getting a speed up of $7.8\times$ and $1.4\times$, respectively.
The underlying reason is that our training benefits more from the affinity mimicking supervision and the good initialization using weight inheritance. This also proves that weight inheritance is capable of accelerating cross-modal distillation.

\begin{table}[t]
\resizebox{0.48\textwidth}{!}{%
\begin{tabular}{cccccc}
\toprule
Model  & IN-V2 & IN-A & IN-R & ObjectNet & IN-Sketch \\
\midrule
CLIP ViT-B/32 \cite{clip}  & 56.0 & 31.6 & 69.4 & 29.9 & 42.3 \\
OpenCLIP ViT-B/32 \cite{open_clip}  & 55.1 & 21.7 & 73.4 & 28.9 & 49.4 \\
\rowcolor{gray!10}
\textbf{TinyCLIP ViT-63M/32(Ours)}  & 55.7 & 22.8 & 74.1 & 31.2 & 50.8 \\
\rowcolor{gray!10}
\textbf{TinyCLIP ViT-45M/32(Ours)}  & 52.6 & 19.8 & 71.5 & 29.1 & 48.8\\
\bottomrule
\end{tabular}
}
\vspace{2mm}
\caption{Zero-shot robustness evaluation. CLIP \cite{clip} is trained on WIT-400M, OpenCLIP \cite{open_clip} and TinyCLIP is trained on LAION-400M \cite{laion400m}. Top-1 accuracy is reported.}
\label{tab:robustness}
\vspace{-3mm}
\end{table}

\subsection{Transfer Learning Results}

\textit{Classification on 23 datasets.} We also evaluate the performance of zero-shot and linear-probe classification on 23 datasets by the toolkit Elevater \cite{elevater}. As shown in Tab.~\ref{tab:zs-ic}, our model wins on 7 datasets in zero-shot benchmark, and 9 datasets on linear-probe benchmark. The performance of our model is close to OpenCLIP ViT-B/32 \cite{open_clip}, where the two models are both trained on Laion-400M \cite{laion400m}. 

\textit{Zero-shot robustness evaluation.} We verify the robustness of our models on 5 datasets, which are out of ImageNet distribution. As shown in Tab. \ref{tab:robustness}, our TinyCLIP ViT-63M/32 outperforms OpenCLIP ViT-B/32 \cite{open_clip}, and wins 3 datasets when compared with CLIP ViT-B/32 \cite{clip}.
Our TinyCLIP ViT-45M/32 uses 50\% fewer parameters yet still wins on IN-R and IN-Sketch, compared to CLIP ViT-B/32. This demonstrates the robustness of the distilled TinyCLIP models and the efficacy of our proposed distillation method.

\vspace{-2mm}
\section{Conclusions}
In this paper, we propose TinyCLIP, a method for distilling large-scale vision-language pre-trained models. It mainly introduced two core techniques named affinity mimicking and weight inheritance. Extensive experiments and ablation studies have demonstrated the effectiveness of TinyCLIP, showing that it can largely reduce model size while maintaining competitive performance. In the future, we will explore ways to further improve cross-modal distillation efficiency on extremely small models. 

\vspace{-2mm}
\section*{Acknowledgement}
Prof. Hongyang Chao was partially supported
by NSFC (U1611461, 61672548) and NSFC (U22A2095). Prof. Xinggang Wang was partly supported by NSFC (No. 62276108).

{\small
\bibliographystyle{ieee_fullname}
\bibliography{egbib}
}

\end{document}